\documentclass[letterpaper, 10 pt, conference]{ieeeconf}  %

\IEEEoverridecommandlockouts                              %

\let\labelindent\relax
\usepackage{graphics} %
\usepackage{epsfig} %
\usepackage{times} %
\usepackage{amsmath} %
\usepackage{amssymb}  %
\usepackage{multirow}
\usepackage{pbox}
\usepackage{amsthm} 
\usepackage{blindtext}
\usepackage{duckuments}
\usepackage{booktabs}
\usepackage[dvipsnames]{xcolor}
\usepackage{tabularx}
\usepackage{boldline, bbm}
\usepackage{color}
\usepackage{enumitem}
\usepackage{multicol}
\usepackage{float}
\usepackage{pifont}
\usepackage{cite}
\usepackage{bbm}
\usepackage{graphicx}
\graphicspath{{./figures/}}
\usepackage{multicol}        %
\usepackage{algorithm}
\usepackage{algpseudocode}
\usepackage[export]{adjustbox}
\makeatletter
\let\NAT@parse\undefined
\makeatother
\usepackage[
colorlinks,
linkcolor=blue,
citecolor=blue,
filecolor=red,
urlcolor=blue]{hyperref}
\usepackage[]{todonotes}

\usepackage{balance}

\newcommand{\PlannerName}{\texttt{MapEx}} %

\newcommand{\xxnote}[3]{}
\ifx\hidenotes\undefined
  \renewcommand{\xxnote}[3]{\color{#2}{#1: #3}}
\fi

\title{\LARGE \bf
MapEx: Indoor Structure Exploration with \\ Probabilistic Information Gain from Global Map Predictions 
}

\algrenewcommand\algorithmicrequire{\textbf{Input:}}
\algrenewcommand\algorithmicensure{\textbf{Output:}}

\author{Cherie Ho$^{*1}$, Seungchan Kim$^{*1}$, Brady Moon$^{1}$, Aditya Parandekar$^{2}$, Narek Harutyunyan$^{3}$, \\ Chen Wang$^{4}$, Katia Sycara$^{1}$, Graeme Best$^{5}$, Sebastian Scherer$^{1}$
\thanks{* Equal Contributions}%
\thanks{This work was supported by the NSF Graduate Research Fellowship under Grant No. DGE1745016, ARL award W911NF2320007, AIST project OSP00011981, and CMU GSA/Provost Conference Funding.}
\thanks{$^{1}$ Carnegie Mellon University Robotics Institute}
\thanks{$^{2}$ BITS, Pilani – Goa \hspace{9mm} $^{3}$ Brown University}%
\thanks{$^{4}$ University at Buffalo \hspace{7.2mm} $^{5}$ University of Technology, Sydney}%
\thanks{Corresponding Authors:\tt{ \{cherieh,seungch2\}@andrew.cmu.edu  }}}
\begin{document}

\maketitle
\thispagestyle{empty}
\pagestyle{empty}

\begin{abstract}

Exploration is a critical challenge in robotics, centered on understanding unknown environments.
In this work, we focus on structured indoor environments, which often exhibit predictable, repeating patterns. Conventional frontier-based exploration approaches have difficulty leveraging this predictability, relying on simple heuristics such as `closest first' for exploration. More recent deep learning-based methods predict unknown regions of the map for information gain computation, but these approaches are often sensitive to the predicted map quality or fail to account for sensor coverage. 
To overcome these issues, our key insight is to jointly reason over what the robot can observe and its uncertainty to calculate \textit{probabilistic information gain}. We introduce $\PlannerName$, a new exploration framework that uses predicted maps to form probabilistic sensor model for information gain estimation. $\PlannerName$ generates multiple predicted maps based on observed information, and takes into consideration both the computed variances of predicted maps and estimated visible area to estimate the information gain of a given viewpoint. Experiments on the real-world KTH dataset showed on average 12.4\% improvement than representative map-prediction based exploration and 25.4\% improvement than nearest frontier approach.\\
Website: \textcolor{blue}{ \href{https://mapex-explorer.github.io/}{https://mapex-explorer.github.io/}}
\end{abstract}

\section{Introduction}
\label{sec:introduction}
Exploration is an important problem in robotics, with applications ranging from indoor search and rescue operations~\cite{sampedro2019fully} to planetary missions~\cite{schilling1996mobile} and outdoor monitoring~\cite{hitz2014fully}. The primary objective of exploration is to plan paths in an unknown environment to maximize understanding of the environment within a limited budget. This objective may involve building an accurate map~\cite{yamauchi1998mobile}, identifying objects of interest~\cite{chaplot2020object}, and finding traversable routes for robot teams~\cite{zhang2022fast}. The key challenge in exploration is choosing how to navigate to maximize information gain, which requires reasoning over unobserved space due to uncertain environment geometry, occlusions, sensor range limits, and sensor noise. 

\begin{figure}[t]
    \centering
\includegraphics[width=\linewidth]{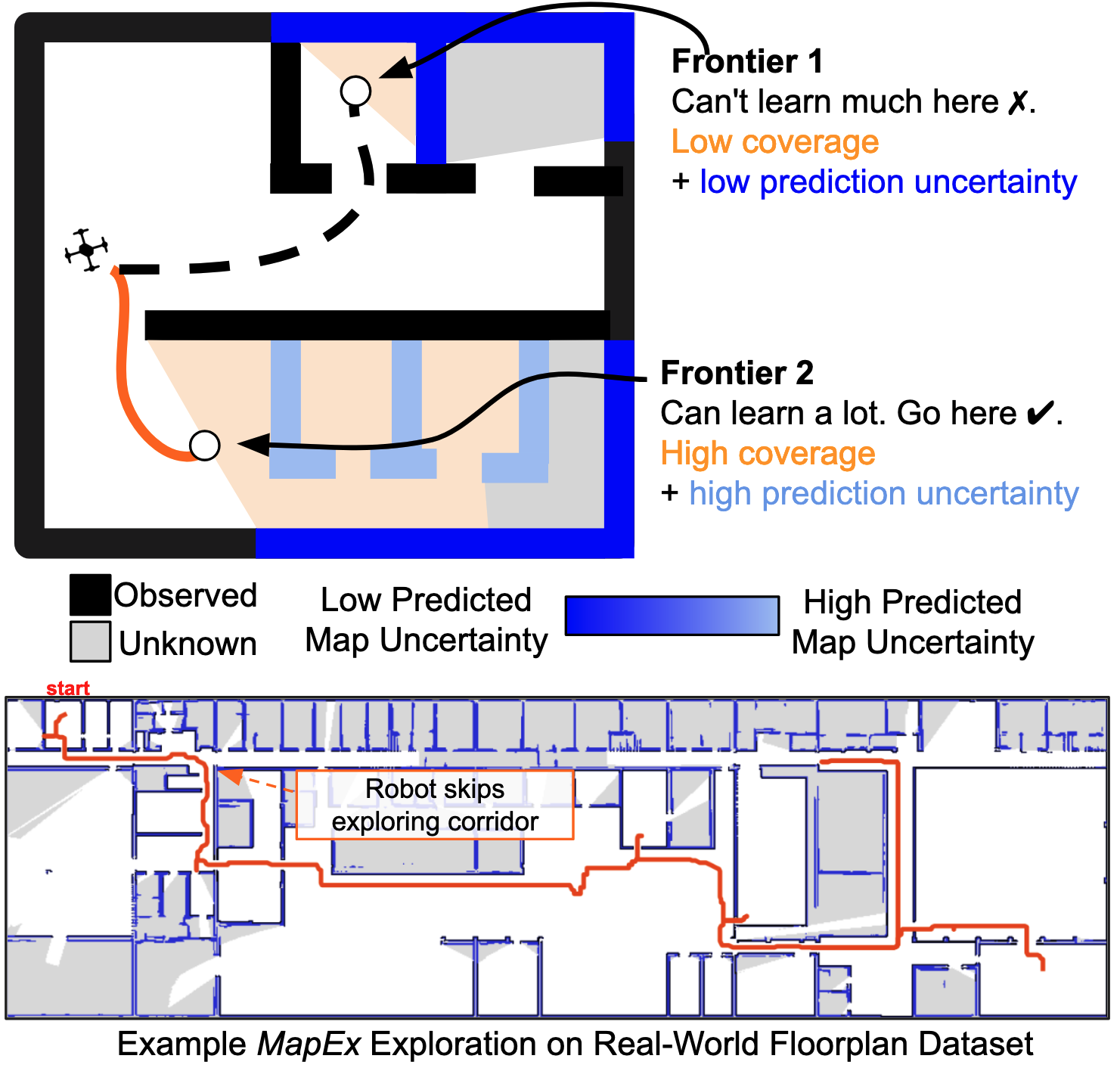}    

\caption{We present \PlannerName, a new framework that uses predicted maps for indoor exploration. Our key insight is to calculate \textit{probabilistic} information gain that \textit{jointly} reasons on coverage and uncertainty. We validate with a real-world floor plan dataset and show superior exploration performance compared to related map-prediction based exploration algorithms.}
\label{intro-figure}
\end{figure}

Many real-world environments possess inherent structure, predictability, and repeatability of environment geometry. For example, indoor environments such as offices or hospitals consist of repeated rooms and corridors, where observing partial views of the space can provide valuable information for predicting the overall layout. However, most existing approaches have difficulty leveraging this predictability for more efficient exploration.
For instance, conventional frontier-based exploration~\cite{yamauchi1997frontier, best2023multi} do not directly reason over the predictable nature of the environment. Instead, they rely on heuristics such as ``visit closest area first". 

To leverage such structural predictability, recent work has explored using deep learning techniques to predict unknown regions of maps, and use such prediction to improve exploration. For instance, IG-Hector \cite{shrestha2019mappred} uses a predicted map to estimate how much area can be observed from a viewpoint. However, such viewpoint scoring heuristic is sensitive to the accuracy of the predicted map. UPEN \cite{georgakis2022upen} instead leverages such uncertain map prediction behavior, by explicitly predicting map uncertainty and then going towards these areas. However, this method does not consider sensor coverage or visibility, which do not reflect actual sensor acquisition. Therefore, our key insight is to jointly reason over what the robot can observe and its uncertainty to form \textit{probabilistic information gain} from multiple predicted maps.

We propose $\PlannerName$, a novel exploration framework that uses predicted maps to form probabilistic sensor model as information gain metric. $\PlannerName$ generates multiple predicted maps from the observed information, from which we compute mean and variance maps. We take the variances of predicted maps and estimated visible area into consideration to compute the information gain of a given viewpoint for exploration planning. We tested $\PlannerName$ with the real-world KTH floor plan dataset\cite{aydemir2012kthdataset}, demonstrating superior exploration performance and topological understanding by 12.4\% to SOTA predicted-map based exploration methods and by 25.4\% to nearest-frontier method.

In summary, our contributions are as follows:
\begin{itemize}
    \item We present a new robot exploration framework that uses predicted maps to form probabilistic sensor model as information gain metric for exploration planning.
    \item We use the information gain metric to augment frontier-based exploration, which achieves improved exploration results on the real-world KTH floor plan dataset~\cite{aydemir2012kthdataset} over SOTA methods that uses map predictions (IG-Hector\cite{shrestha2019mappred}, UPEN\cite{georgakis2022upen}).
    \item Finally, we show the predicted maps produced by $\PlannerName$ have improved global topological understanding and higher utility for downstream path planning.
\end{itemize}

\section{Related Work}
\label{sec:related_work}

\subsection{Robot Exploration}
Exploring an unknown environment, or finding paths to build a map, are core challenges in robotics research. Traditional robotic exploration approaches use the concept of frontiers \cite{yamauchi1997frontier}, the boundaries between known and unknown space. Robots greedily select the next frontiers to visit, by using distance-based measures \cite{yamauchi1997frontier} or variants such as viewpoint selection and graph-search heuristics \cite{best2022resilient}. Another category of popular approaches uses information theory; these approaches \cite{bourgault2002information, bai2016information} seek to maximize the information gain over the next actions. Other approaches include topological \cite{kim2013topological} and graph-based \cite{cao2021tare} approaches that model topological representations of the environment and build graphs that are scalable to large environments. 

\subsection{Image Inpainting}

Image inpainting is a widely-researched topic in Computer Vision that focuses on reconstructing missing parts of an image \cite{bertalmio2003simultaneous, hays2007scene}. 
Recently, with the advent of deep learning, many image inpainting works proposed methods that leverage various architectures such as convolutional neural network \cite{liu2018image}, encoder-decoder \cite{pathak2016context}. 
More recently, image inpainting is extensively used in robot applications to predict maps, such as completing semantic segmentation map\cite{song2018spg, ho2024map} and reconstructing offroad terrain map \cite{stolzle2022reconstructing, triest2024unrealnet, aich2024deepbayesianfuturefusion}. In this work, we use LaMa \cite{LAMA}, a large-mask image inpainting network, to fill in unseen regions and complete a global predicted map given the observation.

\subsection{Map Prediction for Exploration}
To leverage structural predictability, there is a body of work on image inpainting techniques to predict unobserved areas of a map for exploration \cite{chang2007pslam, georgakis2022upen, luperto2022indoorpred, ramakrishnan2020occupancy, tao2024learnexplore, zwecher2022integratingdeep, shrestha2019mappred, ericson2024beyondfrontier, saroya2020topological, sharma2023pre, tan20244cnet}. The field is summarized in a survey paper \cite{tan2022enhancing}. We find work that uses Hough line features of observed maps to predict maps by searching for most similar pre-built maps in a dataset \cite{luperto2022indoorpred}. However, the perception input and output may not be expressive enough for maps with more complex geometry. 

Deep learning emerges as a promising approach for map prediction as the models and outputs are more expressive. Several works leverage deep learning-based map predictors with deep reinforcement learning-based planners for exploration \cite{ramakrishnan2020occupancy, tao2024learnexplore, zwecher2022integratingdeep}. However, these approaches are limited to short horizon decisions, such as 2.5m away from the robot \cite{ramakrishnan2020occupancy}, one step towards a direction \cite{tao2024learnexplore, zwecher2022integratingdeep}. 

Our work builds on successful approaches that leverage deep-learning map prediction techniques and couple them with classical exploration planning techniques that allows more explicit planning over longer horizons, and no need for data to learn planning \cite{shrestha2019mappred, luperto2021exploration, luperto2022indoorpred, ericson2024beyondfrontier, georgakis2022upen}. One method to using the prediction for exploration is to estimate the potential new information from a predicted sensor coverage at a given viewpoint and use it as an information gain metric \cite{shrestha2019mappred, luperto2021exploration, luperto2022indoorpred, ericson2024beyondfrontier}. However, this information gain metric is susceptible to poor map predictions. Alternatively, UPEN \cite{georgakis2022upen} leverages such uncertainty in map prediction, to bias exploration towards high-uncertainty areas. UPEN estimates map uncertainty by predicting an ensemble of maps and using its variance as uncertainty. However, the information metric used is purely variance over the path and does not consider the potential sensor coverage, which is important to encapsulate actual information gain. To address the limitations of past work, $\PlannerName$ jointly reasons over the predicted sensor coverage and the uncertainty of map prediction to form \textit{probabilistic information gain} that is easily used by classical exploration frameworks \cite{yamauchi1997frontier}, to generate long-horizon and informative exploration paths.

\begin{figure*}[ht!]
    \centering
\includegraphics[width=\linewidth]{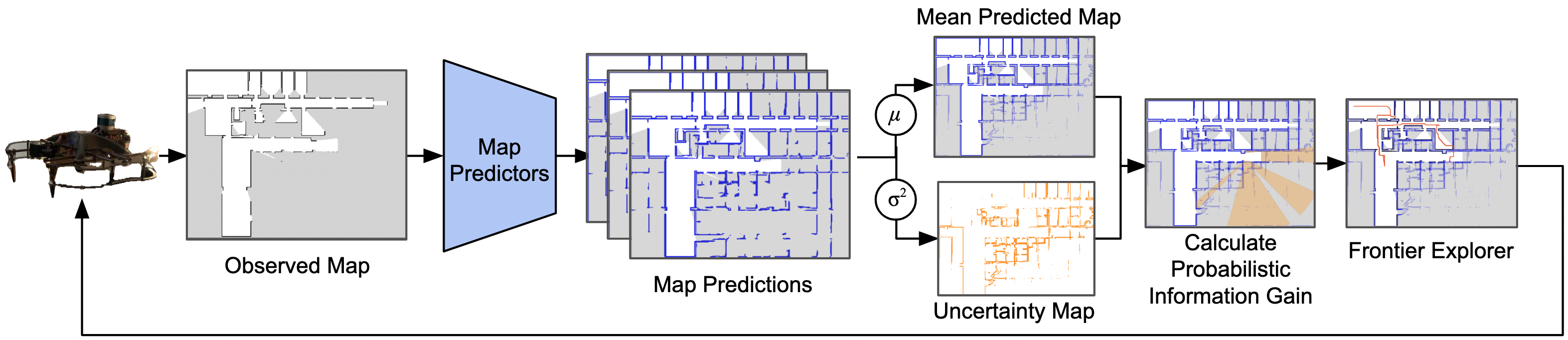}    
\caption{\textbf{\PlannerName~Framework:} $\PlannerName$ estimates \textit{probabilistic information gain} from predicted maps to use in a frontier explorer. A robot accumulates observations into a top-down occupancy map, which is passed to an ensemble of map predictors to generate multiple map predictions. Information gain for each frontier is then estimated based on the mean and variance of the predicted maps. Finally, the frontier with the highest reward is set as the next goal.}
    \label{approach_pipeline}
\end{figure*}

\section{Problem Statement}
We address the problem of robot exploration in unknown indoor environments to gain a comprehensive understanding of the structure. Consider a robot exploring a 2D environment $\mathcal{E}\subset\mathcal{R}^2$. The robot's state is represented as a 2D discrete state $\mathbf{x}=[x, y]$, and the robot's action $a \in \mathcal{A}$ transitions its pose on a diagonally-connected grid at each timestep. The robot is equipped with a 2D LiDAR sensor with range $\lambda$. At time $t=0$, the robot starts with no prior knowledge of the environment, and at each timestep $t$, it receives 2D LiDAR measurements $o_t$ from evenly spaced rays spanning $360^\circ$ originating from the robot state $\mathbf{x_t}$, and takes an action $a_t$ according to the planner. We assume that the robot's pose in a global coordinate frame is known and noise-free LiDAR range measurements are available. It continues to take actions and receive measurements until the time budget $T$.

The robot updates a 2D occupancy grid $O_t$ from a top-down view (or observed map), which represents the accumulation of LiDAR observations $o_{\{1:t\}}$ up until time $t$. An observed map $O_t$ has three labels (0: free, 0.5: unknown, 1: occupied). Along with this, the robot generates 2D predicted map $P_t$ from $O_t$ using a global prediction module; we will explain the details of the global map prediction in Sec.~\ref{sec:approach}.

We aim to plan an exploration path trajectory, or a sequence of states $\{\mathbf{x_0}, \mathbf{x_1} ... \}$, which maximizes understanding of the environment. This can be measured in many different ways, ranging from pixel-wise accuracy to topological usefulness of $O_t$ and $P_t$; we define several relevant metrics to validate our experiments later in Sec.~\ref{sec:results}.

\begin{figure}
    \centering
    
    \includegraphics[width=1\linewidth]{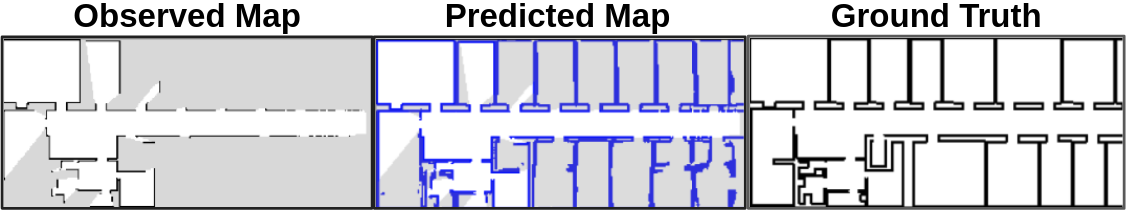}
    
    \caption{\textbf{Example Map Prediction:} Given the observed map, our model predictor can generate a reasonable predicted map that is close to the ground-truth map. These predictions are used in $\PlannerName$ for viewpoint scoring. }
    \label{fig:observed-and-prediction}
\end{figure}

\section{Approach}
\label{sec:approach}

To efficiently explore structured indoor environments, we leverage predicted maps to identify areas with high probabilistic information gain. The $\PlannerName$ pipeline, illustrated in Fig. \ref{approach_pipeline}, consists of four key steps:
\begin{enumerate}
    \item Predicting maps $P_t$ from observed map $O_t$ (Sec.~\ref{global_map_prediction});
    \item Quantifying map prediction uncertainty (Sec.~\ref{uncertainty_estimation});
    \item Estimating probabilistic information gain (Sec.~\ref{proabilistic_raycast_visible_mask});
    \item Using information gain for exploration (Sec.~\ref{MapEx algorithm}).
\end{enumerate}

\subsection{Global Map Prediction}
\label{global_map_prediction}
While LiDAR provides an accurate occupancy map $O_t$ in observed areas, large areas in the environment often remain unobserved due to sensor range and occlusions, especially in indoor settings. Therefore, we use a global map prediction module $\mathcal{G}$, which predicts unknown regions and produces a complete predicted map $P_t$ as %
\begin{equation}
    P_t = \mathcal{G}(O_t).
    \label{eq:global_lama}
\end{equation} 

Given the prevalence of large unknown areas during indoor exploration, we use LaMa network architecture \cite{LAMA}, which has large receptive field and the ability to understand global context, making it suitable for inpainting maps with large unknown areas. We treat a top-down view occupancy grid as a 2D image (where 1 pixel = $0.1$m) and regard the unknown areas of the environment similarly to missing parts of an image. LaMa is used to predict what lies beyond the robot's observation. 
To increase map prediction performance, we fine-tuned LaMa's out-of-the-box Places2 weights using the KTH floor plan dataset\cite{aydemir2012kthdataset}. Our training set consists of \textit{(observed map, ground truth map)} data pairs collected through nearest-frontier exploration. 
Unlike prior works that predicts local maps \cite{shrestha2019mappred, georgakis2022upen}, we generate global predicted maps to enable more comprehensive global planning in exploration. Fig.~\ref{fig:observed-and-prediction} shows an example of an observed map, associated predicted map and ground-truth map.

\subsection{Uncertainty Estimation and Variance Map}
\label{uncertainty_estimation}
While the robot can reasonably predict the map $P_t$ given the observed map $O_t$, there is inherent ambiguity in the unobserved areas. Therefore, the robot should estimate the \textit{uncertainty} of the predictions. To achieve this, we maintain an ensemble of separate LaMa networks $\mathcal{G}_i$, and let each $\mathcal{G}_i$ generate a prediction: 
\begin{equation}
    P_{i,t} = \mathcal{G}_i(O_t).
    \label{eq:lama_ensemble}
\end{equation} 
We used the entire training set to fine-tune $\mathcal{G}$, and split the training set into $n_p$ subsets, each used to fine-tune the parameters of $\mathcal{G}_i$. This allows $n_p$ independent predictions $P_{i,t}$ given the same observed map $O_t$. Using these predictions, we can generate a pixel-wise variance map $V_t$:
\begin{equation}
    V_t = {\textit{Variance}} (P_{i,t}), \quad i \in \{1,2,\dots n_p\}.
    \label{eq:variance_map}
\end{equation}
In this work, we set prediction ensemble size $n_p=3$.

\subsection{Calculating Probabilistic Information Gain}
\label{proabilistic_raycast_visible_mask}

\begin{figure}
    \centering
    \includegraphics[width=1\linewidth]{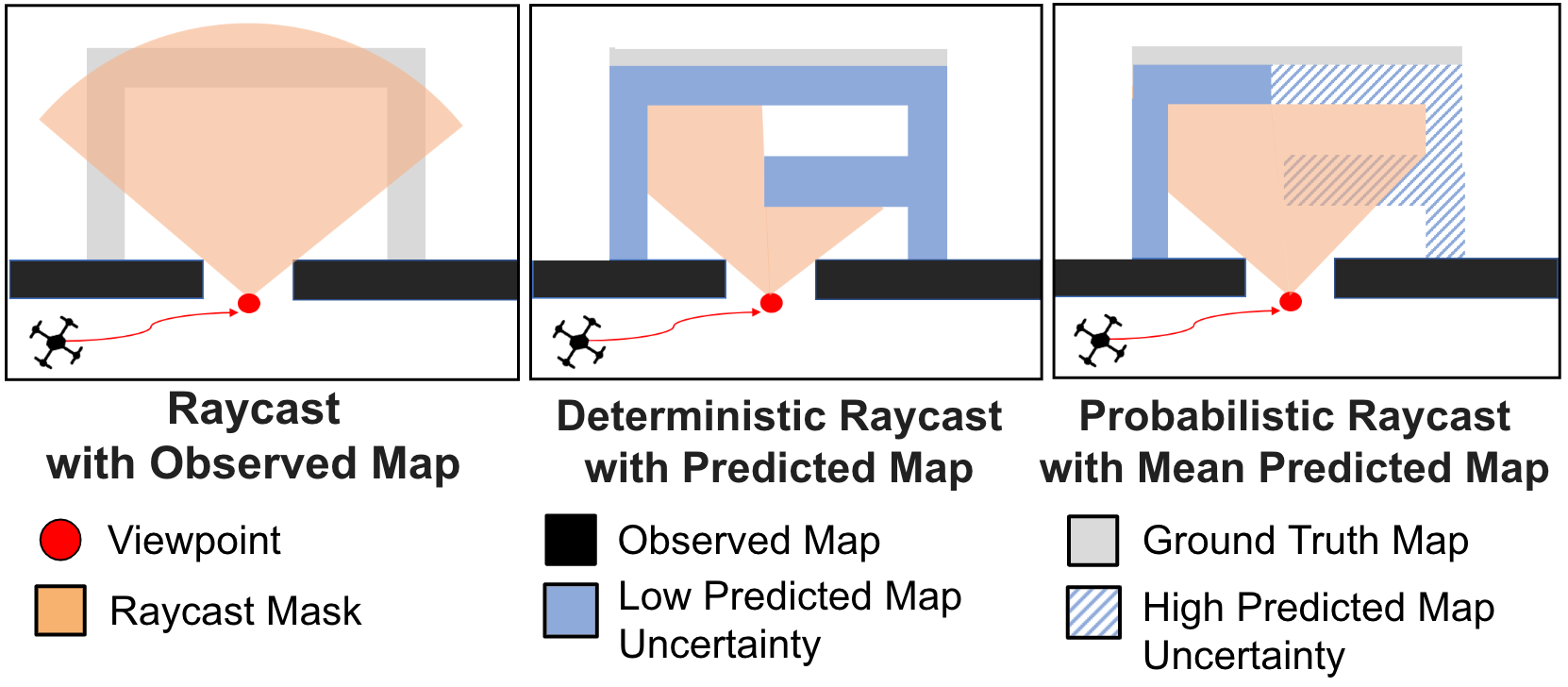}
    \caption{\textbf{Illustration of different raycast methods}: By raycasting only with an observed map, the robot may overestimate the potential information gain. Using a deterministic raycast may lead to incorrect gain estimates when map predictions are not correct. In contrast, $\PlannerName$ uses a probabilistic raycast to reason on potential sensor coverage, resulting in improved exploration.}
    \label{fig:raycast}
\end{figure}

To explore an unknown environment effectively, it's crucial for the robot to move towards areas where it can gain the most information. This requires identifying viewpoints where the sensor coverage is high, as these locations provide the greatest potential for uncovering new areas, and also areas with high uncertainty. Accurate estimation of information gain is essential as poor estimates can lead to inefficient exploration. We detail how we derive \textit{probabilistic information gain} using the predicted and variance maps defined in Sec.~\ref{global_map_prediction} and~\ref{uncertainty_estimation}.
To estimate sensor visibility given uncertain predictions, relying on deterministic raycasting, as done in \cite{luperto2021exploration, luperto2022indoorpred}, can lead to inaccurate results when the predicted maps are unreliable. To address this, we instead employ a probabilistic variant of raycasting. Fig.~\ref{fig:raycast} illustrates the difference between raycast methods.

Given a viewpoint $c \in \mathcal{E}$, we generate $n_l$ hypothetical rays $R_j$ (where $j=1,2,\dots, n_l$) with a range of $\lambda$, emitted in a $360^\circ$ pattern originating from $c$. In a deterministic raycast approach, each ray $R_j$ stops when it hits an occupied cell, and if there are no obstructing cells along its path, $R_j$ reach the end at a distance of $\lambda$. In this work, we use a \textit{probabilistic raycast} approach instead: first, we calculate the mean predicted map $\overline{P_t}$, which is the average of predicted maps $\{P_{i,t}\}$. Each ray $R_j$ starts with an initial accumulated occupancy value $\Delta=0$, and as $R_j$ passes through the pixels of $\overline{P_t}$, it adds the occupancy value of each pixel in the mean map $\overline{P_t}$  to $\Delta$. When $\Delta$ reaches a threshold $\epsilon$, the raycast terminates. 
Given ray end points, we perform a flood fill to generate a sensor coverage mask. We then mask out already observed areas from the sensor coverage mask to generate a visibility mask $\nu$, which probabilistically represents the unknown area estimated to be visible at location $c$. We then compute information gain $\textbf{I}$ by summing the values of all pixels in the variance map $V_t$ that also fall within the visibility mask $\nu$. In other words, 
\begin{equation}
    \textbf{I} = \sum_{(x_k,y_k)} V_t[x_k,y_k], \quad \forall(x_k,y_k) \in \nu.
    \label{eq:info_gain}
\end{equation}
This information gain $\textbf{I}$ estimates the potential reduction in uncertainty, not just by considering the size of observable area, but also the variance it will cover. Through empirical testing, we found that $\epsilon=0.8$ yields the best performance. 

\subsection{$\PlannerName$: Map Exploration with Prediction}
\label{MapEx algorithm}
Here, we explain how the  $\PlannerName$ algorithm is structured and uses the information gain values to augment a frontier-based explorer for efficient exploration. When selecting the next frontier, the robot first generates a predicted map $P_t$ based on the current observed occupancy grid $O_t$. Additionally, it generates an ensemble of predictions $\{P_{i,t}\}$, from which it computes the variance map $V_t$. The frontiers are extracted from the boundary between known and unknown cells in $O_t$, similar in \cite{yamauchi1997frontier}. We cluster frontiers using connected components, filter out small ones, and use the centroids of the remaining clusters as final frontiers. When calculating the score for these frontiers, the information gain is measured as described in Sec.~\ref{proabilistic_raycast_visible_mask}. To discourage backtracking, we divide each frontier score by the Euclidean distance from the current pose of the robot to each frontier, so that if the scores are similar, the robot prefers to select a closer frontier. Once the waypoint is chosen, a local path is generated using an A* local planner, and the robot moves along the path. The pseudocode of the algorithm is provided in Alg.~\ref{alg:MapEx}.

\begin{algorithm}[t]
\caption{$\PlannerName$ Exploration Planner}\label{alg:MapEx}
\begin{algorithmic}[1]
\Require environment $\mathcal{E}$, time budget $T$, start pose $\mathbf{x_0}$ 
\State \textbf{for} $t$ in $T$:
\State \hspace{1.5mm} observation $o_t$ $\leftarrow$ $\mathbf{x_t}$, update occupancy grid $O_t$
\State \hspace{1.5mm} \textbf{if} \text{waypoint $\Psi$ is not available:}
\State \hspace{3mm} prediction $P_t \leftarrow \mathcal{G}(O_t), P_{i,t} \leftarrow \mathcal{G}_i(O_t)$\Comment{Eq.~\ref{eq:global_lama},~\ref{eq:lama_ensemble}}
\State \hspace{3mm} variance map $V_t \leftarrow P_{i,t}$ \Comment{Eq.~\ref{eq:variance_map}}
\State \hspace{3mm} extract frontiers $\mathcal{F} \leftarrow O_t$
\State \hspace{3mm} \textbf{for} frontier $f$ in $\mathcal{F}$:
\State \hspace{4.5mm} visibility mask $\nu \leftarrow \textit{probabilistic raycast}(f, P_{i,t})$
\State \hspace{4.5mm} info gain $I \leftarrow \sum_{(x_k,y_k)\in \nu} V_t(x_k,y_k) $ 
\Comment{Eq.~\ref{eq:info_gain}}
\State \hspace{4.5mm} score frontier $f.score \leftarrow I/||\textbf{x}_t-f||_2$
\State \hspace{3mm} waypoint $\Psi \leftarrow$ choose frontier with maximum score
\State \hspace{1.5mm} $a_t \leftarrow$ \text{local planner($\Psi$)}, transitions to $\textbf{x}_{t+1}$
\Ensure Observed map $O_T$, predicted map  $P_T$
\end{algorithmic}
\end{algorithm}
\begin{figure*}[t]
\centering
\includegraphics[width=1\linewidth]{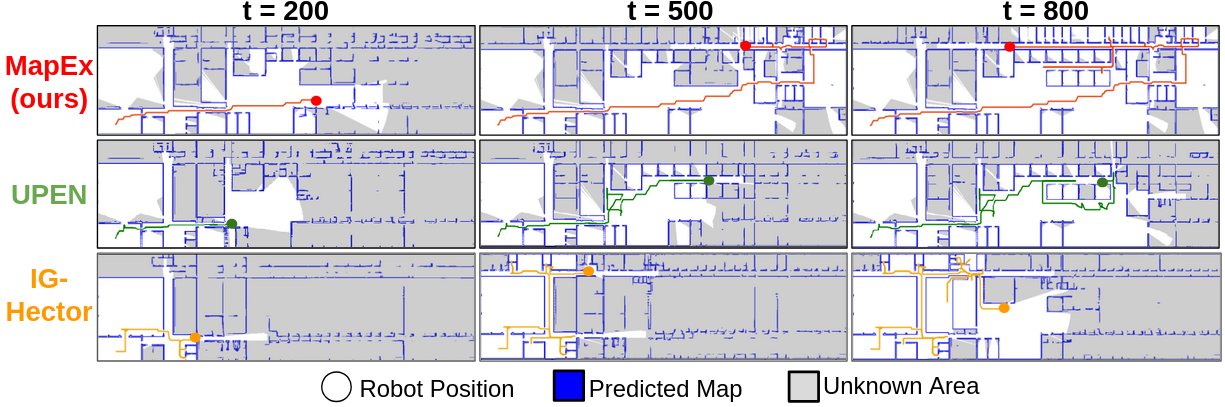}
\caption{\textbf{Exploration Progress over Time: } $\PlannerName$ explores the map best resulting in the most accurate predicted map. In contrast, UPEN \cite{georgakis2022upen} which optimizes for high variance without sensor coverage, searches a limited amount of area. IG-Hector \cite{shrestha2019mappred} only considers sensor coverage and therefore we hypothesize it does not explore sufficiently at the start due to inadequate map predictions. }
\label{fig:exploration_with_time}
\end{figure*}

\section{Experiments}
\label{sec:results}
We evaluate $\PlannerName$ in a simulator using real-world floor plan datasets, comparing its performance against state-of-the-art baselines and conducting ablation studies. 

\subsection{Real-world Floor Plan Dataset}
To benchmark real-world performance, we use the KTH floor plan dataset \cite{aydemir2012kthdataset}, which after removing repeated maps, contains 149 campus floor plan blueprints described as XML files of wall and door locations for each room. We adapt the processing code from \cite{caley2016learnstructure} to obtain the occupancy map and manually correct inaccuracies in the processed map. We then downsample the map to $0.1$m per pixel. 
For training the map predictor and then testing the navigation, we split the floor plans into an 80:20 train-test split. We take care not to include floor plans from the same building into both the train and test set, to test the generalizability of our method. 

\subsection{Experimental Setup}

All methods were tested on 10 held-out floor plans in the KTH dataset, over 1000 timesteps. The robot starts at the 4 corners of each floor plan, resulting in a total of 40 \{floor plan, start pose\} initial conditions. The robot is equipped with a LiDAR with a range of $20$m and 2500 samples per scan. For the map predictor, we fine-tuned with 2367 images in the training set, with a batch size of 4 for 40 epochs.

\subsection{Baselines}
We select three baselines for comparison, from which one is a classical frontier-based approach and two are representative map prediction-based exploration methods.
\begin{itemize}
    \item \textit{Nearest-Frontier}\cite{yamauchi1997frontier}: A classical exploration approach that visits frontiers with lowest euclidean distance. 
    \item \textit{IG-Hector}\cite{shrestha2019mappred}: Map-prediction based  approach that uses the predicted map to estimate a viewpoint's sensor coverage and use it as information gain metric.
    \item \textit{UPEN}\cite{georgakis2022upen}: Map-prediction based  approach that uses an ensemble of map predictions to estimate map prediction uncertainty that is used as exploration heuristic. 
\end{itemize}

We also ablate the impact of different components of the information gain metric used by $\PlannerName$, to quantify the relative impact. For all ablation methods, we kept the frontier exploration framework the same, with only the frontier scoring method different.
Specifically, we investigate
\begin{itemize}
    \item \textit{Deterministic} - Using deterministic raycast, instead of probabilistic raycast, on the predicted map.
    \item \textit{No Variance} - Summing number of pixels, instead of variance, in visibility mask, on the predicted map. 
    \item \textit{Observed Map} - Using observed map for raycast. As there is no predicted maps to produce variance, we sum number of pixels in the visibility mask. 
    \item \textit{No Visibility} - No raycast performed. We sum the variance within $5$m of the viewpoint. 
\end{itemize}

\subsection{Metrics}

We evaluate $\PlannerName$ using a variety of metrics to comprehensively evaluate the information gained by the robot: coverage, predicted map quality, and topological understanding.

\textit{Coverage}: Percentage of the map observed by the robot. 

\textit{Predicted Map Quality}: The intersection over union (IoU) of the predicted 2D occupancy map and the ground truth occupancy map. Specifically, we use the IoU of the \textit{occupied} class within the building footprint. A higher IoU indicates a more accurate predicted map.

\textit{Topological Understanding}: We propose a new metric, Topological Understanding (TU), to directly evaluate how much the outcome of a robot's exploration aids downstream path planning tasks. 
The TU metric is defined as follows: given the exploration result up to timestep $t$, represented by the predicted map $P_t$, we use it to plan an A* path from a start point to a goal. A path is considered successful if it reaches the goal. A path is considered a failure if it collides with an occupied cell in the ground truth map or fails to reach the goal.
We measure success rates for 100 random goal locations per map across different methods.

\begin{figure}[t]
\centering
\includegraphics[width=\linewidth]{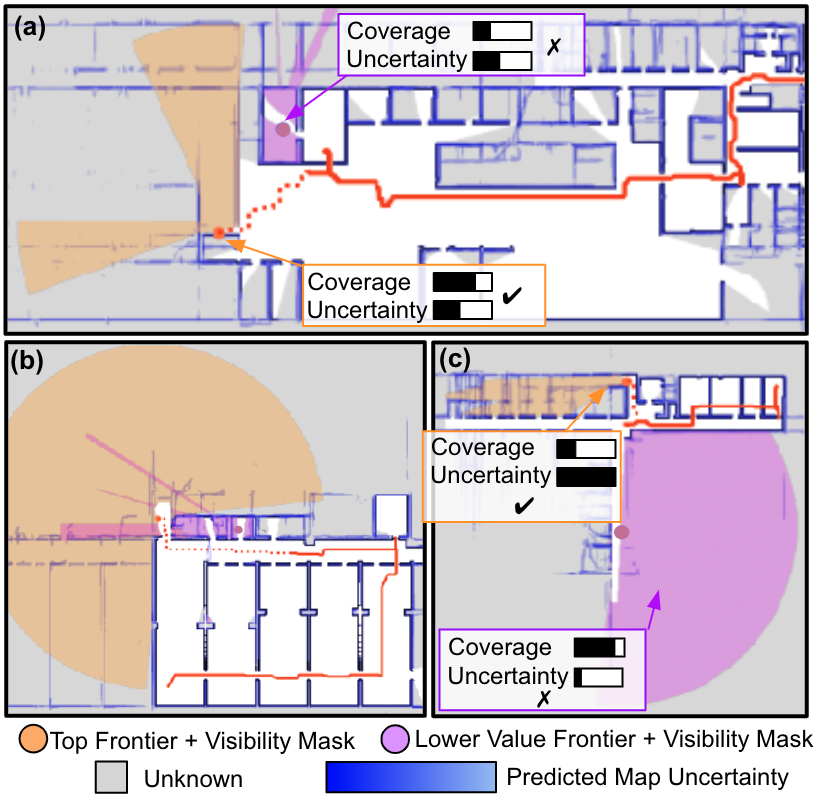}
\caption{\textbf{Examples of $\PlannerName$ frontier scoring: } We show examples of selected frontiers compared to a lower value frontier, to show the use of \textit{jointly} reasoning on sensor coverage and uncertainty to guide exploration.}
\label{fig:frontier_comparisons}
\end{figure}
\subsection{Comparison to Baseline Exploration Methods}
\subsubsection{Qualitative Analysis}
Fig.~\ref{fig:exploration_with_time} compares the exploration progress over time for a test map using different methods. UPEN \cite{georgakis2022upen} reasons on variance at the path and does not consider sensor coverage, likely leading it to explore areas with high variance near the center of the map. In contrast, at $t=500$, $\PlannerName$ also explores the center, but quickly shifts to a more informative hallway, estimating high sensor coverage. IG-Hector \cite{shrestha2019mappred}, which focuses on sensor coverage, struggles early likely due to poor map predictions, causing inefficient exploration of tight spaces. In comparison, as seen in $t=200$,~$\PlannerName$ searches a relatively open area, gathering information about surrounding rooms. Overall, $\PlannerName$ explores more informative regions within the same timeframe, resulting in a higher-quality predicted map compared to the baselines.  Fig.~\ref{fig:frontier_comparisons} provides additional qualitative examples of $\PlannerName$'s frontier scoring, showcasing the advantage of jointly reasoning on sensor coverage and uncertainty. Fig.~\ref{fig:frontier_comparisons}a,b show top-scoring frontiers with high sensor coverage with medium uncertainty. In contrast, Fig. \ref{fig:frontier_comparisons}c presents a case where $\PlannerName$ selects a frontier that covers smaller area but has high uncertainty, rather than one with higher coverage but less potential for learning,  highlighting the importance of uncertainty-aware exploration.

\subsubsection{Quantitative Analysis}
\begin{figure}
    \centering
    \includegraphics[width=1\linewidth]{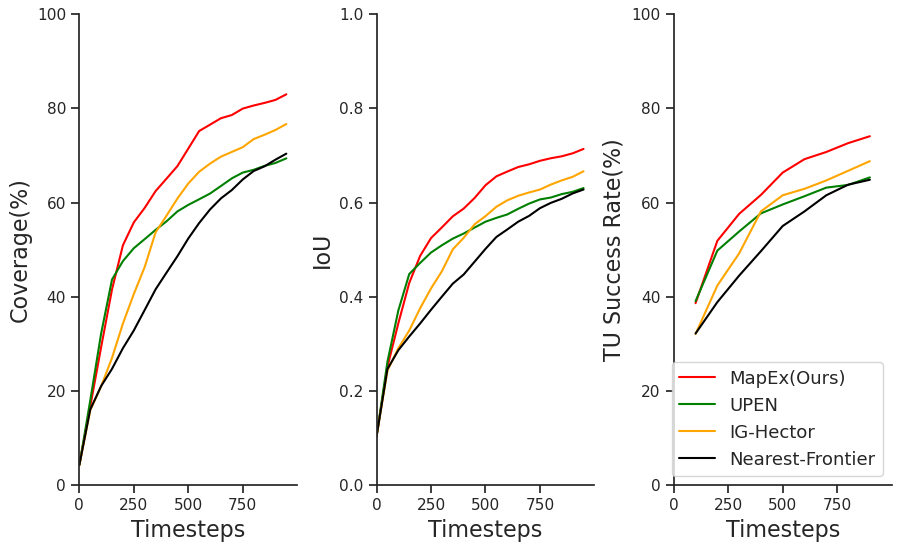}
    \caption{$\PlannerName$ outperforms map prediction-based and nearest exploration planners in terms of Coverage, IoU and Topological Understanding metrics.}
    \label{fig:results_compare_baselines}
\end{figure}

We compare $\PlannerName$'s exploration performance against the baselines in terms of coverage and predicted IoU. As shown in Fig.~\ref{fig:results_compare_baselines}, $\PlannerName$ achieves the best performance. The area under the curves show $\PlannerName$ outperforms \textit{Nearest} by 33.3\% in coverage and by 22.9\% in IoU, while improving on UPEN \cite{georgakis2022upen} by 16.2\% in coverage and 9.9\% in IoU, and on IG-Hector \cite{shrestha2019mappred} by 15.5\% in coverage and 12.3\% in IoU. \textit{Nearest} focuses on close frontiers, resulting in low coverage and IoU. UPEN initially shows steep gains by optimizing for high variance areas, but plateaus as it visits areas of low information gain. IG-Hector prioritizes high sensor coverage but struggles early likely due to poor map prediction. In contrast, $\PlannerName$ reasons about both sensor coverage and uncertainty, yielding superior results throughout. We further compare $\PlannerName$ with baselines using Topological Understanding (TU) metric to quantify the utility of the produced predicted maps (Fig.~\ref{fig:results_compare_baselines}). 
$\PlannerName$ surpasses \textit{Nearest} by 20.1\%, UPEN by 9.6\%, and IG-Hector by 11.1\%. Across all metrics, $\PlannerName$ demonstrates 25.4\% better performance than \textit{Nearest}, and 12.4\% better than map-prediction based exploration baselines. 

\subsection{Ablation Studies}
\begin{figure}
    \centering
    \includegraphics[width=1\linewidth]{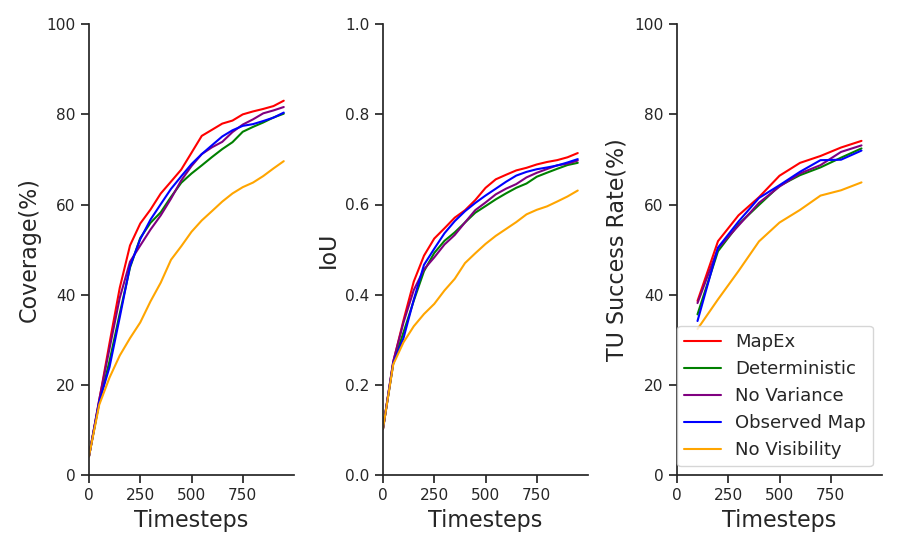}
    \caption{Ablation comparisons of $\PlannerName$ in terms of Coverage, IoU and Topological Understanding metrics.}
    \label{fig:ablations}
\end{figure}

Next, we ablate the inclusion of different components of the information gain metric used by $\PlannerName$ to quantify the relative impact. Fig.~\ref{fig:ablations} compares the coverage, predicted IoU, and topological understanding. We find reasoning on sensor coverage to be critical, as \textit{No Visibility}, which does not use visibility mask and only sums variance in nearby areas, shows poor exploration. Both \textit{No Variance} and \textit{Deterministic} show superior performance by considering sensor coverage. Interestingly,  \textit{Observed Map}, which reasons on observed map, shows competitive performance. This demonstrates the need for a good information gain metric when leveraging predicted maps. By jointly reasoning on sensor coverage and uncertainty, the full $\PlannerName$ exhibits superior exploration performance and produces higher-utility predicted maps.

\section{Conclusion}
\label{sec:conclusion}

In this work, we proposed $\PlannerName$, a novel framework for exploration planning in indoor environments with structural predictability. The core of our approach is a \textit{probabilistic information gain} metric, calculated from multiple predicted maps, which enables the exploration planner to jointly reason about sensor coverage and uncertainty of a given viewpoint. The metric is integrated into a frontier-based exploration framework as a frontier score. Extensive experiments on a real-world floor plan dataset demonstrate that $\PlannerName$ significantly improves exploration performance compared to related map prediction-based algorithms, validating the utility of our frontier scoring method. Finally, we show that the explored maps exhibit superior topological understanding, making them more useful for downstream path planning tasks. For future work, we are interested in extending to multi-robot exploration \cite{kim2023multi} and exploring usage of reinforcement learning in a frontier framework \cite{hu2023opera} for improved performance.

\section{Acknowledgement}
We thank Siddarth Narasimhan for sharing a more processed KTH dataset and providing valuable insights into baseline reproduction.
We additionally thank Simon Stepputtis, Yaqi Xie, John Keller, David Fan and Mononito Goswami for helpful discussions. 
\balance

\bibliographystyle{ieeetr}
\bibliography{references}

\end{document}